# A Novel Approach to Breast Cancer Histopathological Image Classification Using Cross-Colour Space Feature Fusion and Quantum-Classical Stack Ensemble Method

Sambit Mallick[1*], Snigdha Paul[1] and Anindya Sen[1]

[1] Department of Electronics and Communication Engineering,
Heritage Institute of Technology, Kolkata, West Bengal, India
[sambit.mallick.ece25, snigdha.paul.ece25]@heritageit.edu.in
Anindya.sen.heritageit.edu

**Abstract.** Breast cancer classification stands as a pivotal pillar in ensuring timely diagnosis and effective treatment. This study with histopathological images underscores the profound significance of harnessing the synergistic capabilities of colour space ensembling and quantum-classical stacking to elevate the precision of breast cancer classification. By delving into the distinct colour spaces of RGB, HSV and CIE L*u*v, the authors initiated a comprehensive investigation guided by advanced methodologies. Employing the DenseNet121 architecture for feature extraction the authors have capitalized on the robustness of Random Forest, SVM, QSVC, and VQC classifiers. This research encompasses a unique feature fusion technique within the colour space ensemble. This approach not only deepens our comprehension of breast cancer classification but also marks a milestone in personalized medical assessment. The amalgamation of quantum and classical classifiers through stacking emerges as a potent catalyst, effectively mitigating the inherent constraints of individual classifiers, paving a robust path towards more dependable and refined breast cancer identification. Through rigorous experimentation and meticulous analysis, fusion of colour spaces like RGB with HSV and RGB with CIE L*u*v, presents an classification accuracy, nearing the value of unity. This underscores the transformative potential of our approach, where the fusion of diverse colour spaces and the synergy of quantum and classical realms converge to establish a new horizon in medical diagnostics. Thus the implications of this research extend across medical disciplines, offering promising avenues for advancing diagnostic accuracy and treatment efficacy.

# 1 Introduction

## 1.1 Motivation

Breast cancer is a prevalent and challenging health condition characterized by the uncontrolled growth of malignant cells within breast tissue. It remains a critical issue, necessitating the development of highly accurate classification methods. The principal objective of this study showcases the advancements achieved through the cross colour space feature fusion as well as the quantum-classical stacking ensemble approach in the domain of breast cancer classification using histopathological images. Breast cancer poses a significant health challenge, necessitating the development of highly accurate and reliable classification methods. Despite notable progress, there remains a persistent need for further enhancements in accuracy and robustness. However, a lack of comprehensive exploration in two critical areas—colour space utilization and the quantum-classical ensembling techniques—has impeded progress in the field. To address these gaps, this research investigates the potential of colour space ensembling to improve breast cancer classification across diverse tissue types. By combining multiple colour spaces and using quantum classical stacking, the researchers aim to strengthen classification models and enable more precise discrimination.

## 1.2 Background

The literature survey summarizes research in breast cancer classification, colour space analysis, and quantum-classical classifier fusion—crucial for early detection and treatment. By reviewing diverse literature, it highlights progress, trends, and gaps, offering a resource for researchers and practitioners to inspire further investigations in diagnosis and treatment.

Shen, Wu, and Suk [1] explore the application of deep learning techniques in medical image analysis and the potential of convolutional neural networks (CNNs) to improve accuracy in various medical imaging tasks. Das, Kaur and Walia [2] discuss the application of various feature extraction methods, such as texture analysis, shape descriptors, and deep learning-based approaches, to capture distinctive patterns of breast cancer in histopathological images. Anurag et al. [3] focus on defining descriptors with local attention by dividing images into patches, providing a robust representation of breast cancer histopathological images for malignancy identification.

In addition to the mentioned works, the utilization of colour space analysis in breast cancer research has also been explored. Colour space provides valuable information about the characteristics of tissues and lesions, aiding in improved classification and diagnosis. Li and Plataniotis [4] focus on the comparison of RGB, HSV, and LAB colour models and the methodology of using H&E stained images, which are commonly used in histopathology for diagnosis of breast cancer. In the field of breast cancer research, while the utilization of colour space analysis has been explored, the specific method of colour space ensembling has not been extensively investigated. However, in other domains of image analysis, colour space ensembling has shown promising results.

For instance, Xiang and Suandi [5] proposes a technique for extracting human skin colour regions from colour images by combining the RGB and YUV colour spaces.

Recently, the integration of quantum classifiers in breast cancer classification has emerged as a promising research direction. Projapoti et al. [6] investigate the potential benefits and challenges of quantum computing in breast cancer research. Shan et al. [7] demonstrate the feasibility and effectiveness of QSVM for breast cancer detection by applying it to a breast cancer dataset.

In addition to the integration of quantum classifiers, the use of model ensemble stacking methods has gained attention in breast cancer classification. Kumar et al. [8] contributes to the field of breast cancer detection by introducing an optimized stacking ensemble learning model that leverages the strengths of multiple machine learning algorithms. The findings highlight the potential of ensemble learning techniques in improving breast cancer diagnosis and classification. Chatterjee et al. [9] focuses on integrating quantum evolutionary optimization, weighted type-II fuzzy system, staged Pegasos Quantum Support Vector Classifier (QPSVC), and multi-criteria decision-making (MCDM) system to improve the accuracy of breast cancer diagnosis and grading.

- The novelty of this approach lies in utilizing cross-colour space ensemble of an unexplored method in breast cancer classification.
- This research presents a unique opportunity to investigate quantum-classical stacking ensemble methods, offering potential benefits for advancing breast cancer diagnosis and treatment accuracy.

The following Section 2 discusses the datasets used, Section 3 presents the algorithms employed, Section 3 presents the experimental results and finally, the conclusion is presented in the last section, summarizing the key findings and implications of the research.

## 2 Material

### 2.1 Dataset

The BreakHis dataset 400X [10], a histopathological goldmine for biomedical image analysis researchers, provides high-resolution, annotated images of breast cancer tissue samples at 400x magnification as shown in Figure 1. This diverse range of images enables the investigation of tumor morphology, cellularity, and tissue architecture, advancing the understanding of breast cancer pathology and contributing to the development of more accurate diagnostic and prognostic tools. For this research, the authors have utilized a subset of 360 images from this dataset to further explore and analyze breast cancer characteristics.

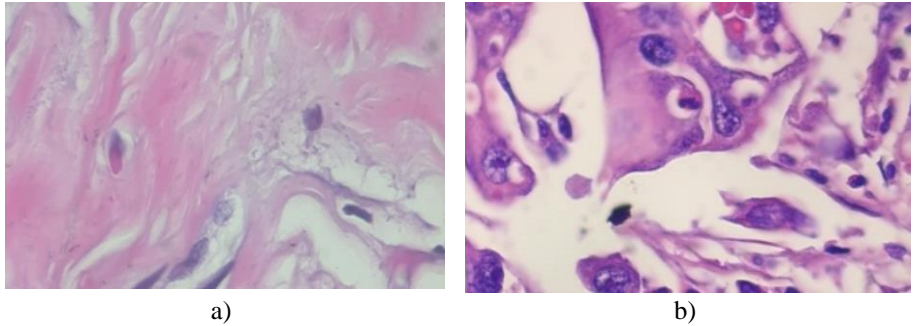

a) b)
**Figure 1.** BreakHis 400X dataset a) Benign cell, b) Malignant cell

### 2.2 Hardware and Software

Our research work involved Python code for model training and inference, implemented using Jupyter Notebooks. We utilized virtual machines provided by Google Colaboratory for running our code. To handle image processing tasks, we heavily relied on two open-source computer vision libraries, namely OpenCV and scikit-image. By leveraging these tools, we were able to ensure efficient execution and effective analysis of our research work, achieving optimal results.

## 3 Proposed Methods

Within the scope of this study, the authors introduced a comprehensive approach aimed at examining the impact of cross-colour space feature ensembling on the BreakHis dataset. The authors also explored the ensembling of quantum and classical models via a stacking method. By conducting a systematic investigation, the authors sought to evaluate the collective influence of these distinct techniques, thus fostering a more comprehensive understanding of their implications for the dataset under scrutiny.

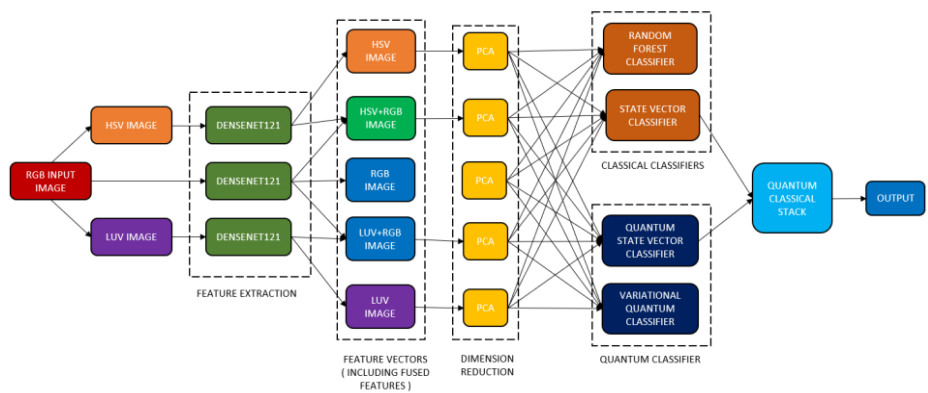

**Figure 2.** Proposed method workflow

Initially, the authors converted the RGB images to HSV and CIE L*u*v. This conversion process allowed for a broader representation of the image data, capturing additional aspects in case of HSV, and perceptual uniformity in the case of CIE L*u*v. The entire procedure outlined in Figure 2 is presented, with a detailed explanation provided below.

### 3.1 RGB → HSV Conversion

HSV colour space represents colours based on hue (H), saturation (S), and value (V). In case of 8-bit and 16-bit images, R, G, and B are converted to the floating-point format and scaled to fit the 0 to 1 range.

$$V \to MAX\{R, G, B\} \text{ -(1)} \qquad S \to \begin{cases} \frac{V - \min\{R,G,B\}}{V} & if\ V \neq 0 \\ 0 & otherwise \end{cases} \quad (2)$$

$$H \to \begin{cases} \frac{60(G-B)}{V-\min\{R,G,B\}} & if\ V=R \\ 120 + \frac{60(B-R)}{V-\min\{R,G,B\}} & if\ V=G \\ 240 + \frac{60(R-G)}{V-\min\{R,G,B\}} & if\ V=B \\ 0 & if\ R=G=B \end{cases} \quad (3)$$

If H<0 then H ← H + 360. On output $0 \le V \le 1, 0 \le S \le 1, 0 \le H \le 360$

- For 8-bit image $V \leftarrow 255V, S \leftarrow 255S, H \leftarrow H/2$

### 3.2 RGB → CIE L*u*v Conversion

It is a colour model that represents colours based on their lightness (L), chroma (u), and chroma (v) components, which represents the perceived brightness of the colour, and the chromaticity or colourfulness of the image. The CIE L*u*v colour space is designed to provide a perceptually uniform representation of colours.

$$\begin{bmatrix} X \\ Y \\ Z \end{bmatrix} \to \begin{bmatrix} 0.412453 & 0.357580 & 0.180423 \\ 0.212671 & 0.715160 & 0.072169 \\ 0.019334 & 0.119193 & 0.950227 \end{bmatrix} \begin{bmatrix} R \\ G \\ B \end{bmatrix} \quad (3)$$

$$L = \begin{cases} 116 * Y^{1/3} - 16, & for\ Y > 0.008856 \\ 903.3Y, & for\ Y \le 0.008856 \end{cases} \quad (5)$$

$$u' \leftarrow 4 * X/(X + 15 * Y + 3Z)$$
$$v' \leftarrow 9 * Y/(X + 15 * Y + 3Z)$$

$$u \leftarrow 13 * L * (u' - u_n) \text{ where } u_n = 0.19793943 \quad (6)$$
$$v \leftarrow 13 * L * (v' - v_n) \text{ where } v_n = 0.46831096 \quad (7)$$

The outputs $0 \le L \le 100, -134 \le u \le 220, -140 \le v \le 122$

- For 8-bit images
  $L \leftarrow 255/100 L, u \leftarrow 255/354(u + 134), v \leftarrow 255/262(v + 140)$

### 3.3 Cross Colour Space Feature Ensemble

In the pursuit of advancing histopathological image analysis, the authors explored a novel technique by ensembling features from RGB, HSV, and CIE L*u*v colour spaces through averaging which aimed to leverage the unique information provided by each colour representation, enhancing the model's performance and robustness. By reducing the dimensionality to 10 features using PCA and ensembling the features, the authors created a unified representation for each image, capturing diverse patterns and improving accuracy. The ensemble technique acted as regularization and facilitating the detection of subtle features in the images for the classical and quantum classifiers.

### 3.4 Classical and Quantum Classification Models

In this study, the authors independently employed classical and quantum classification models. The classical models utilized well-established machine learning algorithms, while the quantum models explored cutting-edge quantum computing principles.

**Support Vector Machine (SVC).**

Support Vector Machine (SVC) [11] is a powerful supervised learning algorithm used for classification and regression tasks. It finds the hyperplane that optimally separates data points of different classes, maximizing the margin between them for better generalization. SVC can handle non-linearly separable data through kernel functions, mapping data into higher-dimensional spaces. The regularization parameter (C) controls the balance between maximizing the margin and minimizing classification errors. SVC's versatility extends to image recognition, text categorization, and medical diagnosis. Its robustness, solid theoretical foundation, and adaptability have made it a widely used and effective tool in various domains of machine learning.

**Random Forest Classifier.**

The Random Forest Classifier [12] is a potent machine learning algorithm designed specifically for classification tasks. It constructs multiple decision tree models during training and combines their predictions through voting to determine the final class label. By building diverse trees on different subsets of data and features, it reduces overfitting and enhances the model's ability to generalize to new, unseen data. This approach allows Random Forest to effectively handle complex datasets with multiple features and class imbalances, making it a popular and powerful choice for various classification.

**Quantum Support Vector Machine (QSVC).**

The Quantum Support Vector Classifier (QSVC) [13] is a leading quantum machine learning algorithm. It distinguishes itself by harnessing the kernel trick to map classical data into a high-dimensional quantum feature space. This technique empowers QSVC to achieve precise and efficient classification, making it a favored choice in quantum machine learning and holding promise for advancements in diverse applications.

The quantum counterpart to the Classical SVM is defined by selecting a feature map:

$$\Psi: \mathbb{R}^r \to S(2^q) \tag{8}$$

$$x \to |\psi(x)\rangle\langle\psi(x)| \tag{9}$$

where, $S(2^q)$ is the space of density matrix on q qubits and kernel function is given by:

$$k(x, y) = tr[|\psi(y)\rangle\langle\psi(y)||\psi(x)\rangle\langle\psi(x)|] \tag{10}$$

The QSVC is implemented using the ZZFeatureMap for the data encoding part. It is used for 2 repetitions in the circuit with a linear entanglement of qubits. The classification rule can be restated in the familiar SVM form:

$$\widetilde{m}(x) = sign\left(2^{-n} \sum_a w_a(\vec{\theta}) + \phi_a(\vec{x}) + b\right) \tag{11}$$

To find the optimal $w_a$, one can employ kernel methods and the standard Wolfe - dual of the SVM can be taken under consideration.

**Variational Quantum Classifier (VQC).**

The Variational Quantum Classifier (VQC) [14] represents a widely adopted supervised quantum-classical hybrid technique, particularly applicable in NISQ devices and simulators. This approach employs iterative measurements to calculate the cost function, enabling error mitigation opportunities. By utilizing variational algorithms and differential programming, classical data is mapped into an expanded quantum feature space. The process involves state preparation encoding classical data sets into qubit amplitudes and rotations for quantum hardware or simulators. Parameterized unitary operations executed on qubits are adjustable following predefined rules. Ultimately, the quantum execution produces the output, effectively categorizing the input data.

### 3.5 Quantum-Classical Stack Ensemble

In this groundbreaking study aimed at enhancing breast cancer malignancy detection, the authors ingeniously ensembled the classical and quantum classification models using stacking. Leveraging the power of conventional machine learning techniques and the emerging capabilities of quantum computing, this hybrid approach showcased a novel strategy in medical diagnostics to unlock new opportunities for improved accuracy and efficiency in breast cancer diagnosis.

**Stack Ensemble.**

The stacking ensemble [15] [16] involves training a classical machine learning model and a quantum machine learning model separately on the same dataset. The predictions from both models are used as input features for a meta-classifier, typically a classical machine learning model. The meta-classifier learns to combine the predictions of the classical and quantum models, leveraging their diverse strengths to make the final prediction as shown in Figure 3.

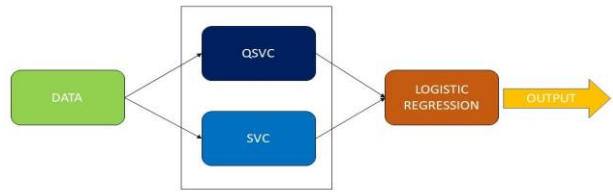

**Figure 3.** Quantum Classical Ensembling Stacking

By stacking a quantum model with a classical model, this ensemble approach aims to harness the advantages of both higher testing accuracy and improved performance.

### 3.6    Evaluation Metric

**Testing Accuracy.**

The authors used testing accuracy as an evaluation metric to assess the performance of their models. By comparing the proportion of correctly classified instances in the test dataset, they gained insights into the effectiveness of the stacked ensemble, combining both the classical and quantum classification models. This evaluation allowed them to validate the model's ability to make accurate predictions on unseen data and assess its overall performance in solving the breast cancer malignancy detection task.

## 4    Results and Discussion

The primary objective of this research is to showcase the advancements achieved through the colour space ensemble method. The authors employed RGB, HSV and CIE L*u*v colour space images to conduct feature extraction using DenseNet121. Subsequently, the authors ensembled different colour space combinations. The resulting graph in Figure 4 exemplifies the notable increase in classification accuracy brought about by the colour space ensemble technique when applied to different classifiers.

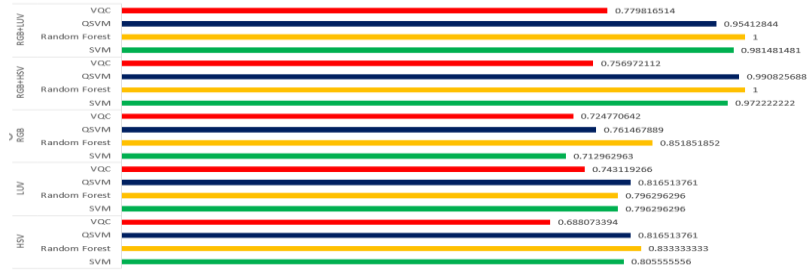

**Figure 4.** Testing Accuracy of Classical and Quantum Classifiers with Colour Space Ensemble

As illustrated in Figure 4, among the classifiers, Random Forest exhibits the highest testing accuracy in case of single colour space, followed by QSVM, SVM, and VQC. However, both RGB+HSV and RGB+CIE L*u*v ensembles outperform the individual colour spaces. Notably, Random Forest achieves a perfect accuracy of 1.0 in both RGB+HSV and RGB+ CIE L*u*v ensembled models, and QSVM achieves 0.99 and 0.95, respectively, for RGB+HSV and RGB+ CIE L*u*v. SVM and VQC also attain relatively higher accuracy when used in conjunction with the ensembled colour spaces. The improvement in classification accuracy can be attributed to the unique features captured by each colour space. HSV colour space emphasizes colour information, CIE L*u*v highlights brightness and contrast, and RGB represents the most common colour representation. By combining RGB with HSV and CIE L*u*v, the ensembled models benefit from a comprehensive representation of the image data, allowing them to detect subtle patterns and characteristics of breast cancer with increased precision. Moreover, ensembling integrates information from multiple colour spaces, thus enhancing the model's ability to maintain accuracy in different scenarios.

Another objective of this research is to explore the ensemble of quantum and classical classifiers to further enhance breast cancer classification accuracy. This experiment is conducted with a focus on the classifiers that performed relatively poorly, namely SVM and VQC. Table 1 presents the comparison of results before and after the stacking ensemble, demonstrating the significant improvements achieved through the fusion.

**Table 1.** Testing accuracy of Quantum-Classical Stack Ensemble

| Model Name | RGB and HSV | RGB and CIE L*u*v |
|---|---|---|
| SVC | 0.972222 | 0.981481 |
| VQC | 0.756972 | 0.779817 |
| SVC+VQC | 1 | 0.954128 |

Individually, SVM achieved testing accuracies of 0.9722 for RGB+HSV and 0.9815 for RGB+ CIE L*u*v. Similarly, VQC achieved testing accuracies of 0.7569 for RGB+HSV and 0.7798 for RGB+ CIE L*u*v, indicating that VQC may not be as effective in capturing the complex patterns and features of breast cancer in these colour spaces. These lower accuracies could be attributed to the limitations and biases inherent in VQC, which might not be well-suited for breast cancer classification in these particular colour spaces. However, when the SVM and VQC classifiers were combined through the stacking ensemble (SVM+VQC), the accuracy significantly improved. The ensemble model achieved a perfect accuracy of 1 for RGB+HSV. Similarly, for RGB+ CIE L*u*v, the stacking ensemble achieved a testing accuracy of 0.9541, which is a substantial improvement over the individual performances of SVM and VQC.

Thus, these approaches effectively leverage the complementary strengths of both classifiers. SVM is known for its ability to handle complex decision boundaries and capture intricate patterns, while VQC can exploit quantum advantages for certain classification tasks. The ensemble method reduces the negative impact of misclassifications that may have occurred when using SVM and VQC individually, leading to an overall enhancement in breast cancer classification accuracy.

## 5 Conclusion

In conclusion, this research presents a novel approach that demonstrates the efficacy of colour space ensembling in enhancing breast cancer classification accuracy by leveraging RGB, HSV, and CIE L*u*v colour spaces. The stacking ensemble of quantum and classical classifiers further improves classification results by compensating for individual classifier limitations, ensuring more reliable breast cancer diagnoses. Future research directions include exploring advanced quantum classifiers and algorithms to achieve higher accuracies, investigating colour space ensembling with other machine learning models like neural networks, and testing on larger and diverse datasets to assess generalizability. Analyzing different colour space combinations and their relevance to specific breast cancer subtypes is crucial, as breast cancer is heterogeneous with distinct characteristics. This research contributes to robust and accurate breast cancer classification systems, facilitating early detection and effective treatment. The integration of colour space ensembling and quantum-classical stacking showcases the potential of combining classical and quantum computing techniques to address complex medical challenges, offering promising prospects for personalized breast cancer diagnosis and treatment strategies.

**Acknowledgement.**

The authors would like to express their heartfelt gratitude to all the faculties of the Department of Electronics and Communication Engineering, Heritage Institute of Technology.

**Disclosure of Interests.**

The authors have no conflicts of interest, financial or otherwise.